\documentclass[11pt,a4paper]{article}
\usepackage[hyperref]{emnlp2020}
\usepackage{times}
\usepackage{latexsym}
\usepackage{graphicx}

\usepackage{caption}
\usepackage{subcaption}

\newcommand\blfootnote[1]{%
  \begingroup
  \renewcommand\thefootnote{}\footnote{#1}%
  \addtocounter{footnote}{-1}%
  \endgroup
}

\usepackage{microtype}

\aclfinalcopy 

\title{HUMAN: Hierarchical Universal Modular ANnotator}

\author{Moritz Wolf$^{1\ast}$, Dana Ruiter$^{12\ast}$, Ashwin Geet D'Sa$^{3\ast}$\\
\textbf{Liane Reiners$^{4}$, Jan Alexandersson$^{1}$, Dietrich Klakow$^{2}$} \\
  $^{1}$DFKI GmbH, $^{2}$Spoken Language System Group, Saarland University \\ 
  $^{3}$Universit\'{e} de Lorraine, CNRS, Inria, LORIA \\
  $^{4}$Department of Communication, Johannes Gutenberg University Mainz \\
  \texttt{\{moritz.wolf, jan.alexandersson\}@dfki.de} \\
  \texttt{\{druiter, dietrich.klakow\}@lsv.uni-saarland.de} \\
  \texttt{ashwin-geet.dsa@loria.fr, liane.reiners@uni-mainz.de}
  }

\date{}

\begin{document}
\maketitle
\begin{abstract}

A lot of real-world phenomena are complex and cannot be captured by single task annotations. This causes a need for 
\emph{subsequent} 
annotations, with interdependent questions and answers describing the nature of the subject at hand.
Even in the case a phenomenon is easily captured by a single task, the high specialisation of most annotation tools can result in having to switch to another tool if the task only slightly changes.
We introduce HUMAN, a novel web-based annotation tool that addresses the above problems by a) covering a variety of annotation tasks on both textual and image data, and b) the usage of an internal deterministic state machine, allowing the researcher to chain different annotation tasks in an interdependent manner.
Further, the modular nature of the tool makes it easy to define new annotation tasks and integrate machine learning algorithms e.g., for active learning.
HUMAN comes with an easy-to-use graphical user interface that simplifies the annotation task and management.
\end{abstract}

\section{Introduction}
\blfootnote{$^{\ast}$ Equal contribution.}
Access to suitable annotated data constitutes a fundamental prerequisite for R\&D of machine learning algorithms and models. 
The demand for new annotated data is growing as new data is collected or existing data is being re-annotated from a new perspective.
In the area of natural language processing (NLP) alone, there is a large variety of tasks that require different types of annotations to be covered by a tool. This includes named-entity recognition or part-of-speech tagging, which require a tool to cover sequence labeling \citep{kiesel-etal-2017-wat}, co-reference and dependency parsing requiring relational annotations \citep{stenetorp-etal-2012-brat, eckart-de-castilho-etal-2016-webAnno, shindo2018pdfanno}, or any type of text classification task requiring document-level annotations \citep{doccano}.

\citet{neves2019review} find that enabling document-level classification is the top missing feature in a large number of reviewed annotation tools. At the same time, document-level annotation has been the most frequently sought-after annotation type according to a recent online survey \citep{survey-annotation-platform}.
Another feature missing from about half of the reviewed tools is multi-label annotations.

Since most annotation tools cover only one or two annotation types, a change in the annotation task can easily require a change in the annotation tool itself. Commercial platforms such as LightTag\footnote{\url{https://www.lighttag.io}} or Prodigy\footnote{\url{https://prodi.gy/}} cover a larger array of tasks to choose from. However, none of these are able to chain a mixture of different tasks (e.g., document-level classification followed by finer-grained sequence labeling) to be performed on a single annotation instance.
One tool that comes close to achieving this, is Angrist\footnote{\url{https://github.com/Tarlanc/angrist}}, however, its lack of modularity makes it difficult to adapt to new annotation scenarios.

Hierarchical Universal Modular ANnotator (HUMAN) follows a highly modular concept, which makes it easy to adapt to a specific annotation scenario.
It uses an internal deterministic state machine (DSM) to guide the annotator through the pre-defined annotation task(s). 
This usage of a DSM allows annotation tasks to be chained in any order needed and makes it easy to implement entirely new annotation tasks and custom features in the future. This \textbf{modular} nature is especially useful when single task annotations do not capture the reality of a problem or when several dependencies exist in the annotations. One example being hate speech corpora \citep{zampieri2019semeval, stru-etal-2019-germeval}, where the target of hate is only supposed to be annotated if a comment has been previously annotated as hateful. 

Further, HUMAN covers a variety of annotation tasks, including the often lacking multi-label annotations and document-level classification, but also sequence labeling on textual data as well as image labeling; a pursuit towards \textbf{universality}. An example involving both document-level annotation as well as sequence labeling is multi-lingual Named Entity Recognition (NER), where the annotator has to identify the language on the document level and then annotate the named entities on the sequence-level. Moreover, when the annotation need is single-task, the fact that many tasks are covered by HUMAN makes it easy to re-use previous installations of HUMAN for a new scenario, even if the task at hand changes. 

Lastly, HUMAN makes the annotation of \textbf{hierarchical} data possible. That is, if an annotation instance is embedded in a context of previous content (e.g., comments in a forum) this context can be shown to the annotators.

The remainder of this paper is organized as follows. In Section \ref{s:system_description}, we explain the structure of HUMAN, starting with the architecture and followed by the internal deterministic state machine, annotation protocol, API, database, and graphical user interface. In Section \ref{s:use_case}, we demonstrate the application of HUMAN for a real-life use case. This is followed by a discussion (Section \ref{s:discussion}) and conclusion (Section \ref{s:conclusion}).

\section{System Description}
\label{s:system_description}

\begin{figure}[t]
    \centering
    \includegraphics[width=.9\columnwidth]{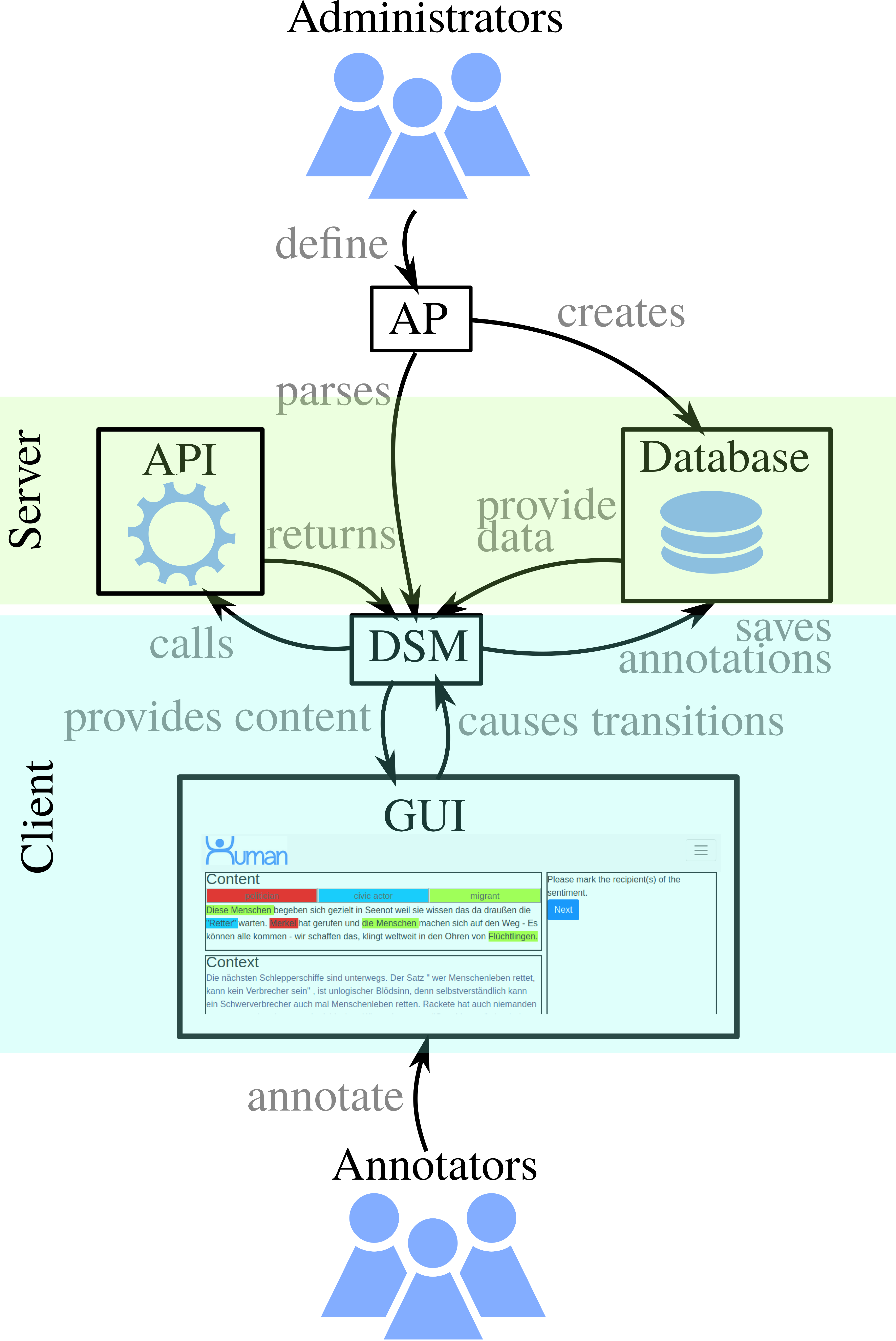}
    \caption{The basic structure behind HUMAN: Administrators define the annotation protocol (AP) which gets parsed to a DSM. The different components in client (GUI and DSM) and server (API and database) interact with each other. Annotators annotate using the GUI.}
    \label{fig:clientserver}
\end{figure}

The HUMAN annotation tool is primarily designed to run on a web server. As such its architecture follows a basic client-server model (Figure \ref{fig:clientserver}). 
Clients and servers exchange messages in a request–response pattern, where the client sends a request to which the server responds.

The \textbf{server}, consisting of the database and the API, serves the code for the client. The database (Section \ref{s:database}) is used for sending new annotation instances to the client or saving finished annotations which are sent by the client.

The \textbf{client} is controlled by a DSM (Section \ref{s:dsm}) to  show an annotation task in the GUI (Section \ref{s:gui}). For this it requests new content and sends finished annotations to the server when an annotation instance is completed. The annotators interact with the GUI to solve annotation tasks.

During \textbf{setup} of the HUMAN system, administrators design the \emph{annotation protocol} (AP), which is a JSON-style definition of the annotation task(s) at hand. This is then used to generate both the database and the DSM.
realize active learning or similar tasks.

The server is implemented using Flask \citep{grinberg2018flask}, a common web framework for Python. The client is written in Typescript and transpiled to JavaScript. The client was tested on Chrome/Chromium (v85.0) and Firefox (v80.0.1).

\subsection{Deterministic State Machine}
\label{s:dsm}

\begin{figure}
    \centering
    \includegraphics[width=\columnwidth]{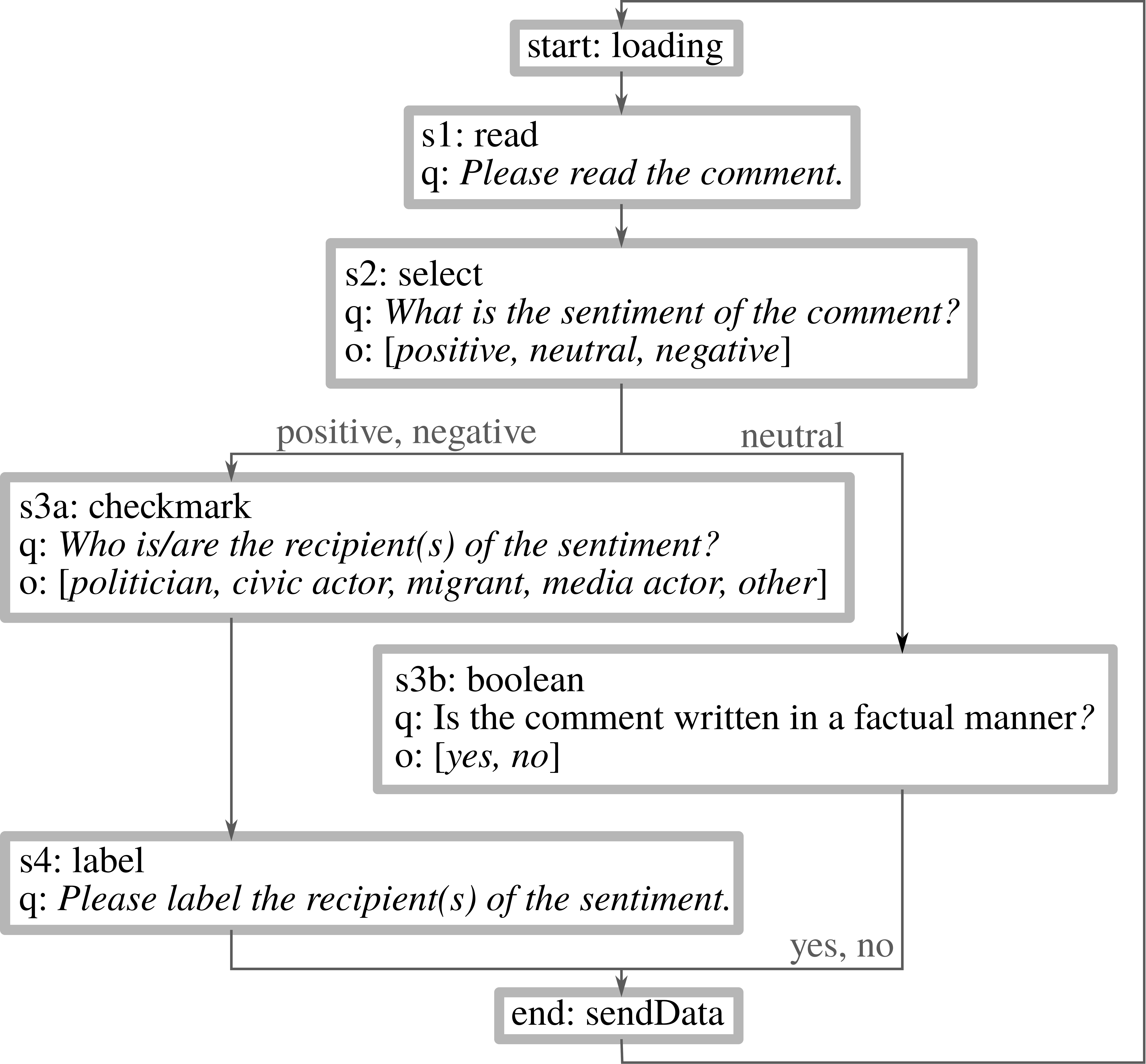}
    \caption{Visualization of a deterministic state machine as used by HUMAN.}
    \label{fig:dsm}
\end{figure}

The back-bone of HUMAN is its deterministic state machine implemented using XState\footnote{\url{https://github.com/davidkpiano/xstate}}, a library for finite state machines and statecharts for JavaScript. It controls how each annotation instance is handled during the annotation process. 

The DSM (Figure \ref{fig:dsm}) starts at a \texttt{start} state which assigns and loads an annotation instance to be presented to the annotator. It then passes through all of the \textbf{annotation states} (AS) that have been defined by the administrator. This allows for the definition of complex \textbf{transitions} between ASs, as the answer an annotator provided to a given question can influence to which subsequent AS the DSM will guide them. As such, it is easy to design a flow of questions presented to the annotator that contains sub-branches and even loops. While accounting for the transitions between ASs, the DSM can also accommodate different \textbf{actions}, such as saving annotations.

By default the DSM comes with three obligatory states:

\begin{itemize}
\item \textbf{start}: Assigns and loads an annotation instance to the annotator.
\item \textbf{failure}: If any unexpected errors occur during annotation, the platform automatically generates an error message and displays it to the user. This state serves as a dead state.
\item \textbf{end}: The end state passes the annotations collected
to the \texttt{annotations} table in the data base.
\end{itemize}

All other states that handle the transitions between questions, need to be defined in the \emph{Annotation Protocol} by the administrator when setting up the HUMAN server.

\subsection{Annotation Protocol}
\label{s:annotation_protocol}

The annotation protocol is the definition of the DSM, using a simplified JSON-style syntax. Within the AP, the project administrators define each annotation task, i.e., state, that should be passed by the DSM. Each state comes with at least two obligatory \textbf{fields}. These are the \emph{transition} field, which describes to which state the DSM should move next, and the state \emph{type}. Depending on the state type, there may be additional fields that further define the quality of a state. The predefined \textbf{state types} are:

\begin{itemize}

\item Functional States
\begin{description}
    \item[loading]: This is usually used to define the type of the \texttt{start} state and loads a textual annotation instance.
    \item[loadingFile]: Analogous to \texttt{loading}, but used to load PDF files or images.
    \item[callAPI]: This state is for calling API functions (Section \ref{s:api}) on the server. It requires the \texttt{api\_call} field.

\end{description}

   \item Annotation States: An annotation state requires at least the additional \texttt{question} field in which an instruction or question to be presented to the annotator is defined.
\begin{itemize}
    \item \textbf{read}: Shows the annotation instance to the annotator.
    \item \textbf{select}: Shows a question and a number of options to choose from. Only one option may be selected by the annotator. \\
    It also requires the \texttt{options} field, which lists the options the annotator can choose from.
    \item \textbf{checkmark}: Analogous to \texttt{select} but allows the choice of multiple options.
    \item \textbf{label}: Prompts the annotator to highlight portions of a text and label it with previously chosen labels.
    \item \textbf{boolean}: Used for yes-no questions.
    \item \textbf{choosePage}: Allows the annotator to choose a page from a PDF to annotate.
    \item \textbf{bbox}: Asks the annotator to set bounding boxes on an image. 
    By writing a custom API call, users can connect models that pre-select parts of an image with bounding boxes.
    \item \textbf{bboxLabel}: Analogous to \texttt{bbox}, but annotators are required to add a text label to each bounding box they place.
\end{itemize}
\end{itemize}

All of these annotation types can be chained after each other in a modular-fashion. State types \texttt{select}, and \texttt{checkmark} can be especially useful for document-level annotations, e.g., creating labels for text classification models. Additionally, \texttt{label} states can be used for sequence annotations such as part-of-speech or named entity tagging, or highlighting certain entities of interest in a text. States \texttt{choosePage, bbox} and \texttt{bboxLabel} can be applied to PDFs or images in which something should be selected. This is practical when creating data for tasks such as optical character recognition or object detection and labeling.

\paragraph{Actions} Each state can be asked to perform different actions. While the \texttt{load} action is handled automatically by the \texttt{loading} type, the \texttt{save} action can be used in ASs and needs to be explicitly stated. When the \texttt{save} action is included in the state definition, then the answer provided by the annotator at this AS will be saved. Non-saving states might be useful to handle transitions in the DSM that are needed to design specific sub-branches, but are not needed during the further assessment of the data annotations.

\begin{figure}
    \centering
    \includegraphics[width=\columnwidth]{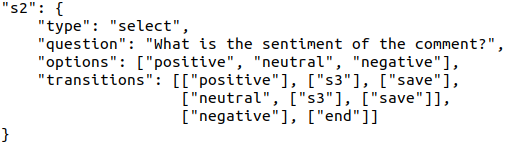}
    \caption{AP definition of a state of type \texttt{select}.}
    \label{fig:ap_example}
\end{figure}

\paragraph{Syntax} The JSON-style AP syntax is simplistic and can best be explained by the example code in Figure \ref{fig:ap_example}, which is the description of a state named \texttt{s2} of type \texttt{select}. It asks the annotator about \emph{the sentiment of the comment} and provides three options \emph{positive, neutral, negative} to the annotator. If the annotator chooses \emph{positive} or \emph{neutral}, the answer will be saved and the DSM directs the annotator to another state named \texttt{s3}. If, however, the annotator chooses \emph{negative}, they are redirected to the \texttt{end} state without saving.

We provide full documentation with instructions on how to define an AP using the defined syntax.\footnote{\url{https://github.com/uds-lsv/human/wiki}}

\paragraph{Parsing} Once the AP is defined, the \texttt{ap\_parser} will parse it into \texttt{XState} compatible format. If any definitions in the AP are ambiguous or undefined, an error message appears that helps the user resolve the problem.

Apart from creating the DSM, the parser also initializes the database instance.

\subsection{API}
\label{s:api}

We provide the possibility to define a custom API on the server. Functions of this API can be called via the DSM with the \texttt{callAPI} state or by adding a \texttt{callAPI} option to a state in the AP. The arguments can contain annotations and/or the current annotation instance.
The reasoning behind this is to enable active learning or similar tasks and have direct access to any machine learning algorithm.

\paragraph{Example}
We want to train an optical character recognition algorithm and need to annotate pictures with bounding boxes around words. We want to show the predicted boxes of the algorithm to the annotator and let them correct them. For that we can write an API function to call the prediction function of our algorithm. In the AP we define a state \texttt{bbox} with the \texttt{callAPI} option and the name of our API function. When entering this state, the DSM will then call the API function with the current annotation instance (in this case a picture). The returned bounding boxes will then be displayed on the picture and corrected by the user.

\subsection{Managing the Database}
\label{s:database}

HUMAN uses an SQLite database with four different tables: \texttt{data}, \texttt{annotations}, \texttt{users}, and \texttt{options}.

\paragraph{Inputs}

Data can dynamically be input into the \texttt{data} table via the GUI. All inputs must be formatted as tab-separated CSV files consisting of the three columns \texttt{content, context, meta}. In \texttt{content}, the content of an instance to be annotated is placed, the \texttt{context} is for optional context information relevant for the annotator, and \texttt{meta} is an encoded JSON object containing any meta information that should be stored with an instance, but that should not be shown to the annotators (e.g., author of a comment, project internal ID of an instance or date of publication).

After uploading the tab-separated CSV file to the server, it will be checked for errors and then parsed into the \texttt{data} table of the database, ready to be distributed to annotators.

\paragraph{Outputs}

At any given time, collected annotations can be downloaded from the \texttt{annotations} table. It is returned as an Excel file.
By default, the name of a column is tied to the name of the state in the DSM that generated the annotation contained. That is, a state named \texttt{s2} will by default write answers into column \texttt{s2}. 
The file also contains the unique instance ID and user ID to match each annotation with its corresponding annotation instance in \texttt{data} and annotator in \texttt{users}.
It is also possible to separately extract each table of the database on the server as tab-separated CSV files.

\paragraph{Users}
Each annotator needs a user account to access the HUMAN server. User information such as username, e-mail, full name and the hashed password are stored in the \texttt{users} table. Further, the user type is stored, in order to separate annotator accounts from administrator accounts. While annotators only have access to the annotation page and their profile information, administrators also have access to the data upload and download page as well as the administrator console.

\paragraph{Options} 
The \texttt{options} table contains information about the set-up of the tool, such as the number of annotators an annotation instance should be assigned to.

\subsection{Graphical User Interface}
\label{s:gui}

\begin{figure}[t]
     \centering
     \begin{subfigure}{\columnwidth}
         \includegraphics[width=\columnwidth]{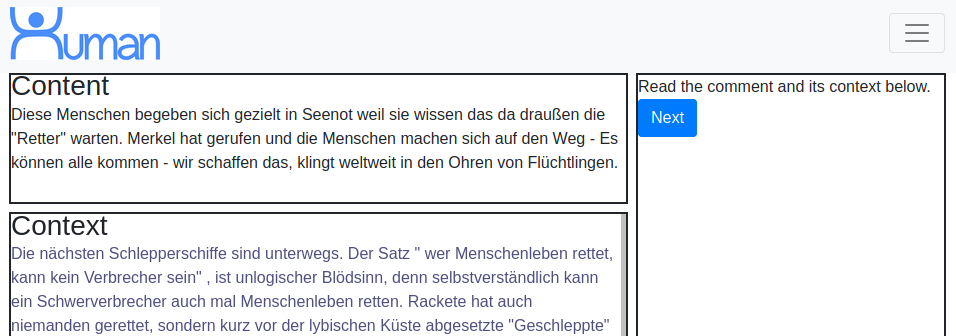}
         \caption{Read question.}
         \label{fig:read_ex}
     \end{subfigure}
     \begin{subfigure}{\columnwidth}
         \centering
         \includegraphics[width=\columnwidth]{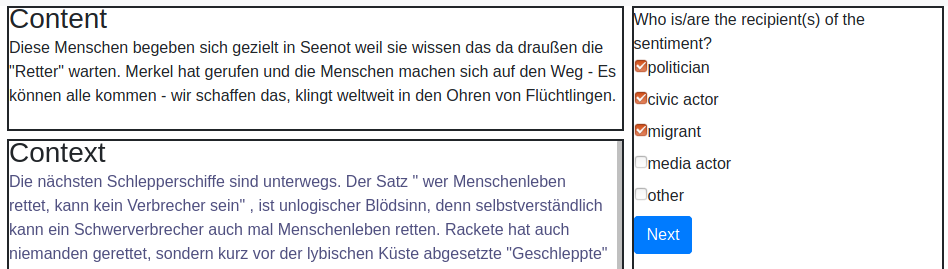}
         \caption{Checkmark question.}
         \label{fig:checkmark_ex}
     \end{subfigure}
    \begin{subfigure}{\columnwidth}
         \centering
         \includegraphics[width=\columnwidth]{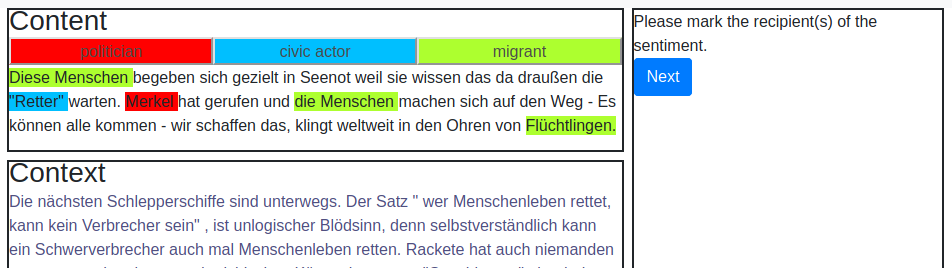}
         \caption{Label question.}
         \label{fig:label_ex}
     \end{subfigure}
        \caption{The annotation page while passing through different annotation states.}
        \label{fig:annotation_page}
\end{figure}

The graphical user interface (GUI) makes the tool easy to use for annotators and administrators. On the \textbf{annotation page} (Figure \ref{fig:annotation_page}), the left-hand side shows the current annotation instance, with content to be annotated on top and optional context information at the bottom. This makes it possible to perform hierarchical annotations, by showing an annotation instance together with the context it was originally embedded in. The right-hand side is dedicated to showing the questions one-by-one as annotators pass through the DSM.

The \textbf{administrator console} in the GUI allows the administrators to manage annotators and annotation settings. This includes activating users that have registered for an account, or deactivating users that should be removed, or changing the annotator's password when required. The administrator can check the total number of annotations made by the annotators.  It is further possible to specify how many annotators are required per annotation instance.
Data upload and download is done via the \textbf{data console} in the GUI and is also only available to administrators.

\section{Use-Case}
\label{s:use_case}
In the case of the interdisciplinary project M-PHASIS, where computer and communication scientists are collaborating to study online hate speech in user-generated content, an annotation tool was needed that can implement multiple facets of hate speech in the AP. 
Annotating in communication science means following a complex AP which consists of different theoretically deduced categories to answer previously defined research questions or hypotheses \citep{frueh-2017}. Often there are multiple levels of analysis units which are hierarchical \citep{roessler-2017}, e.g., news articles with corresponding comments which can be also divided into various statements. Due to the analytical splitting of comments into different statements, it was crucial to have a tool that allows answering specific questions more than once for the same comment (\emph{loops}). 

Due to collaboration across disciplines it was necessary to combine the different types of select and label questions (Section \ref{s:annotation_protocol}). To study media content, communication scholars apply an empirical method called quantitative content analysis, by which they try to systematically categorize content and formal characteristics of messages \citep{frueh-2017}. 
The computational linguists, on the other hand, also required sequence labels to identify specific entities in the texts, e.g., targets of hate speech.

For M-PHASIS there are different questions which depend on the type of text input: news articles (3 select questions, 1 label question, 1 boolean question) and comments. For comments, the questions vary according to type: moderating comment (1 select question) or user comment (10 label questions, up to 30 questions of another type). For the user comments, there are six different thematic blocks in the AP. Some questions must be answered for all comments, other (follow-up) questions must only be answered when certain conditions apply to the particular comment (\emph{branching}). This includes, inter alia, the question of whether statements within a comment contain a positive or negative evaluation of a target or an action recommendation of how to deal with a target, for example when the adaption of a specific behavior is demanded or a threat of physical violence is expressed. Depending on the annotator’s decision, follow-up questions are shown, e.g., regarding the characteristics of a negative evaluation. If required, the tool loops through a block of questions as many times as needed, for example when various evaluations of different targets are expressed in one comment. 

The flexible structure of HUMAN allows one to change the extent of the tool in consideration of the actual content. This enables complex annotations when necessary, but also makes it possible to shorten the AP and therefore the time spent per annotation – something no other tool has been able to accomplish until now.

\section{Discussion}
\label{s:discussion}

HUMAN strongly follows its concept of modularity and allows for the design and implementation of complex annotation protocols. And while it is currently already able to handle a variety of tasks on textual data as well as PDFs and images, many tasks are still uncovered. Two examples here being relationship annotations or asking open answer questions. In order to truly reach \emph{universality}, we envision that the modular nature of the code will invite anyone interested to add new and custom features and annotation types to this open-source tool.

\section{Conclusion and Future Work}
\label{s:conclusion}

We have described HUMAN, a \textbf{modular} annotation tool that covers a variety of annotation tasks, ranging from document-level annotation over sequence labeling to image annotations. Its usage of a deterministic state machine, also accommodates different annotation tasks to be chained in such a way that annotation decisions of the annotator can be followed by different subsequent questions (branching) or the revisions of previous questions (loops). 
Its context and content fields make it possible to perform \textbf{hierarchical} annotations, i.e., annotating an instance together with the context it was embedded in.

This is, as far as we know, the only annotation tool capable of covering such complex annotation needs. This is of use not only for disciplines that require multi-task annotation protocols, but also for various single-task scenarios where users do not want change the tool every time they have a new annotation need with a slightly different task.

While HUMAN is already fully functional and has been used for a real-life annotation scenario, it is a work in progress. Possible new annotation tasks could be e.g., annotations of relationships as in  Brat \citep{stenetorp-etal-2012-brat}, of wave signals, similar to Praat \citep{praat2001} or even videos as in NOVA \citep{Heimerl2019}.

In order to improve accessibility of the tool in the future, we plan to implement a drag-and-drop GUI for the creation of the AP, as well as a visualization of the internally generated DSM to improve debugging. Automatic calculation of statistics such as the inter annotator agreement and  average time spent on an annotation instance are planned.

To further ease the database management, 
administrators should have direct insight on each annotation instance in the database, which can then be added, removed or edited in the GUI without the need of \texttt{SQLite} commands on the server.

The code\footnote{\url{https://github.com/uds-lsv/human}} is published under a GPL-3 licence together with a Wiki with detailed instructions on how to setup the server and define an AP. It also explains how to write custom annotation states and API calls. Two functioning demos of the HUMAN annotation page on two different APs are published on our homepage\footnote{\url{http://human.lsv.uni-saarland.de/}}.

\section*{Acknowledgments}
The development of this tool is partially funded by ANR-DFG Project M-PHASIS (WI 4204/3-1).
A special thanks for all the feedback to Thomas Kleinbauer, Christian Schemer, Laura Ascone, Angeliki Monnier, Irina Illina and Dominique Fohr.

\bibliographystyle{acl_natbib}
\bibliography{anthology,emnlp2020}

\begin{thebibliography}{14}
\expandafter\ifx\csname natexlab\endcsname\relax\def\natexlab#1{#1}\fi

\bibitem[{Boersma and Weenink(2001)}]{praat2001}
Paul Boersma and David Weenink. 2001.
\newblock Praat, a system for doing phonetics by computer.
\newblock \emph{Glot International}, 5(9/10):341--345.

\bibitem[{Eckart~de Castilho et~al.(2016)Eckart~de Castilho,
  M{\'u}jdricza-Maydt, Yimam, Hartmann, Gurevych, Frank, and
  Biemann}]{eckart-de-castilho-etal-2016-webAnno}
Richard Eckart~de Castilho, {\'E}va M{\'u}jdricza-Maydt, Seid~Muhie Yimam,
  Silvana Hartmann, Iryna Gurevych, Anette Frank, and Chris Biemann. 2016.
\newblock \href {https://www.aclweb.org/anthology/W16-4011} {A web-based tool
  for the integrated annotation of semantic and syntactic structures}.
\newblock In \emph{Proceedings of the Workshop on Language Technology Resources
  and Tools for Digital Humanities ({LT}4{DH})}, pages 76--84, Osaka, Japan.
  The COLING 2016 Organizing Committee.

\bibitem[{Fr{\"u}h(2017)}]{frueh-2017}
Werner Fr{\"u}h. 2017.
\newblock \emph{Inhaltsanalyse. Theorie und Praxis}.
\newblock UVK, Konstanz.

\bibitem[{Grinberg(2018)}]{grinberg2018flask}
Miguel Grinberg. 2018.
\newblock \emph{Flask web development: developing web applications with
  python}.
\newblock " O'Reilly Media, Inc.".

\bibitem[{Heimerl et~al.(2019)Heimerl, Baur, Lingenfelser, Wagner, and
  Andr{\'e}}]{Heimerl2019}
Alexander Heimerl, Tobias Baur, Florian Lingenfelser, Johannes Wagner, and
  Elisabeth Andr{\'e}. 2019.
\newblock \href {https://doi.org/10.1109/ACII.2019.8925519} {Nova - a tool for
  explainable cooperative machine learning}.
\newblock In \emph{2019 8th International Conference on Affective Computing and
  Intelligent Interaction (ACII)}, pages 109--115.

\bibitem[{Kiesel et~al.(2017)Kiesel, Wachsmuth, Al-Khatib, and
  Stein}]{kiesel-etal-2017-wat}
Johannes Kiesel, Henning Wachsmuth, Khalid Al-Khatib, and Benno Stein. 2017.
\newblock \href {https://www.aclweb.org/anthology/E17-3004} {{WAT}-{SL}: A
  customizable web annotation tool for segment labeling}.
\newblock In \emph{Proceedings of the Software Demonstrations of the 15th
  Conference of the {E}uropean Chapter of the Association for Computational
  Linguistics}, pages 13--16, Valencia, Spain. Association for Computational
  Linguistics.

\bibitem[{Nakayama et~al.(2018)Nakayama, Kubo, Kamura, Taniguchi, and
  Liang}]{doccano}
Hiroki Nakayama, Takahiro Kubo, Junya Kamura, Yasufumi Taniguchi, and Xu~Liang.
  2018.
\newblock \href {https://github.com/doccano/doccano} {{doccano}: Text
  annotation tool for human}.
\newblock Software available from https://github.com/doccano/doccano.

\bibitem[{Neves and Ševa(2019)}]{neves2019review}
Mariana Neves and Jurica Ševa. 2019.
\newblock \href {https://doi.org/10.1093/bib/bbz130} {{An extensive review of
  tools for manual annotation of documents}}.
\newblock \emph{Briefings in Bioinformatics}.
\newblock Bbz130.

\bibitem[{R{\"o}ssler(2017)}]{roessler-2017}
Patrick R{\"o}ssler. 2017.
\newblock \emph{Inhaltsanalyse}.
\newblock UVK, Konstanz.

\bibitem[{Shindo et~al.(2018)Shindo, Munesada, and
  Matsumoto}]{shindo2018pdfanno}
Hiroyuki Shindo, Yohei Munesada, and Yuji Matsumoto. 2018.
\newblock {PDFAnno: a Web-based Linguistic Annotation Tool for PDF Documents}.
\newblock In \emph{Proceedings of the Eleventh International Conference on
  Language Resources and Evaluation (LREC 2018)}, pages 1082--1086, Miyazaki,
  Japan. European Language Resources Association (ELRA).

\bibitem[{Stenetorp et~al.(2012)Stenetorp, Pyysalo, Topi{\'c}, Ohta, Ananiadou,
  and Tsujii}]{stenetorp-etal-2012-brat}
Pontus Stenetorp, Sampo Pyysalo, Goran Topi{\'c}, Tomoko Ohta, Sophia
  Ananiadou, and Jun{'}ichi Tsujii. 2012.
\newblock \href {https://www.aclweb.org/anthology/E12-2021} {brat: a web-based
  tool for {NLP}-assisted text annotation}.
\newblock In \emph{Proceedings of the Demonstrations at the 13th Conference of
  the {E}uropean Chapter of the Association for Computational Linguistics},
  pages 102--107, Avignon, France. Association for Computational Linguistics.

\bibitem[{Struß et~al.(2019)Struß, Siegel, Ruppenhofer, Wiegand, and
  Klenner}]{stru-etal-2019-germeval}
Julia~Maria Struß, Melanie Siegel, Josep Ruppenhofer, Michael Wiegand, and
  Manfred Klenner. 2019.
\newblock Overview of germeval task 2, 2019 shared task on the identification
  of offensive language.
\newblock In \emph{Proceedings of the 15th Conference on Natural Language
  Processing (KONVENS 2019)}, pages 354--365, Erlangen, Germany. German Society
  for Computational Linguistics \& Language Technology.

\bibitem[{Tan(2020)}]{survey-annotation-platform}
Liling Tan. 2020.
\newblock A survey of nlp annotation platforms.
\newblock https://github.com/alvations/annotate-questionnaire.

\bibitem[{Zampieri et~al.(2019)Zampieri, Malmasi, Nakov, Rosenthal, Farra, and
  Kumar}]{zampieri2019semeval}
Marcos Zampieri, Shervin Malmasi, Preslav Nakov, Sara Rosenthal, Noura Farra,
  and Ritesh Kumar. 2019.
\newblock Semeval-2019 task 6: Identifying and categorizing offensive language
  in social media (offenseval).
\newblock In \emph{Proceedings of the 13th International Workshop on Semantic
  Evaluation}, pages 75--86.

\end{thebibliography}

\appendix

\end{document}